\documentclass[10pt,twocolumn,letterpaper]{article}

\usepackage{cvpr}
\usepackage{times}
\usepackage{epsfig}
\usepackage{graphicx}
\usepackage{graphics}
\usepackage{subcaption}
\usepackage{amsmath}
\usepackage{amssymb}
\usepackage{multirow}
\usepackage[table,xcdraw]{xcolor}
\usepackage{color}
\usepackage{soul}
\usepackage{float}
\usepackage{mathtools}
\usepackage{cuted}

\newcommand{\bx}{\mathbf{x}}

\newcommand{\by}{\mathbf{y}}
\newcommand{\bY}{\mathbf{Y}}

\newcommand{\bR}{\mathbf{R}}
\newcommand{\bK}{\mathbf{K}}
\newcommand{\bp}{\mathbf{p}}
\newcommand{\bc}{\mathbf{c}}
\newcommand{\bk}{\mathbf{k}}
\newcommand{\bW}{\mathbf{W}}
\newcommand{\be}{\mathbf{e}}
\newcommand{\bxi}{\mathbf{\xi}}
\newcommand{\mG}{\mathcal{G}}

\usepackage[pagebackref=true,breaklinks=true,letterpaper=true,colorlinks,bookmarks=false]{hyperref}

\cvprfinalcopy 

\def\cvprPaperID{8930} 

\ifcvprfinal\fi
\begin{document}

\title{MonoPair: Monocular 3D Object Detection Using Pairwise Spatial Relationships}

\author{
    Yongjian Chen \ \ \ \
    Lei Tai \ \ \ \
    Kai Sun \ \ \ \
    Mingyang Li\\
    Alibaba Group\\
{\tt\small \{yongjian.cyj, tailei.tl, sk157164, mingyangli\}@alibaba-inc.com}
}

\maketitle


\begin{abstract}

Monocular 3D object detection is an essential component in autonomous driving while challenging to solve, especially for those occluded samples which are only partially visible.
Most detectors consider each 3D object as an independent training target,
inevitably resulting in a lack of useful information for occluded samples.
To this end, we propose a novel method to improve the monocular 3D object detection by considering the relationship of paired samples.
This allows us to encode spatial constraints for
partially-occluded objects from their adjacent neighbors.
Specifically, the proposed detector computes uncertainty-aware predictions for object locations and 3D distances for the adjacent object pairs,
which are subsequently jointly optimized by nonlinear least squares.
Finally, the one-stage uncertainty-aware prediction structure and the post-optimization module are dedicatedly integrated for ensuring the run-time efficiency.
Experiments demonstrate that our method yields the best performance on KITTI 3D detection benchmark, by outperforming state-of-the-art competitors by wide margins, especially for the  \textit{hard} samples.
\end{abstract}


\section{Introduction}
\label{sec:intro}
3D object detection plays an essential role in  various computer vision applications such as autonomous driving, unmanned aircrafts, robotic manipulation, and augmented reality.
In this paper,
we tackle this problem by using a monocular camera, primarily for autonomous driving use cases.
Most existing methods on 3D object detection require accurate depth information, which can be obtained from either 3D
LiDARs \cite{chen_multiview_2016, qi_frustum_2017, shi_pointrcnn_2018, shin_roarnet_2018, liang_deep_2018, zhou_voxelnet_2017} or
multi-camera systems \cite{chen_3d_2015, chen_3d_2018, li_stereo_2019, pham_robust_2017, qin_triangulation_2019, xu_multilevel_2018}.
Due to the lack of directly computable depth information,
3D object detection using a monocular camera is generally considered a much more challenging problem than using LiDARs or multi-camera systems.
Despite the difficulties in computer vision algorithm design,
solutions relying on a monocular camera can potentially allow for low-cost, low-power, and deployment-flexible systems in real applications. Therefore, there is a growing trend on performing monocular 3D object detection in research community in recent years \cite{brazil_m3d_rpn_2019, chen_monocular_2016, manhardt_roi_10d_2018, mousavian_3d_2017, qin_monogrnet_2018, simonelli_disentangling_2019}.


Existing monocular 3D object detection methods have achieved considerable high accuracy for normal objects in autonomous driving. However, in real scenarios, there are a large number of objects that are under heavy occlusions, which pose significant algorithmic challenges.
Unlike objects in the foreground which are fully visible, useful information for occluded objects is naturally limited.
Straightforward methods on solving this problem are to design networks to exploit useful information as much as possible, which however only lead to limited improvement.
Inspired by image captioning methods which seek to use scene graph and object relationships \cite{dai2017detecting, Li_2017_ICCV, Yao_2018_ECCV} , we propose to fully leverage the spatial relationship between close-by objects instead of
individually focusing on information-constrained occluded objects.
This is well aligned with human's intuition that
human beings can naturally infer positions of the occluded cars from their neighbors on busy streets.

Mathematically, our key idea is to optimize the predicted 3D locations of objects guided by their uncertainty-aware spatial constraints.
Specifically, we propose a novel detector to jointly compute object locations and spatial constraints between matched object pairs.
The pairwise spatial constraint is modeled as a keypoint located in the geometric center between two neighboring objects, which effectively encodes all necessary geometric information. By doing that,
it enables the network to capture the geometric context among objects explicitly.
During the prediction, we impose aleatoric uncertainty into the baseline 3D object detector to model the noise of the output.
The uncertainty is learned in an unsupervised manner, which is able to enhance the network robustness properties significantly.
Finally, we formulate the predicted 3D locations as well as their pairwise spatial constraints into a nonlinear least squares problem to optimize the locations with a graph optimization framework. The computed uncertainties are used to weight each term in the cost function.
Experiments on challenging KITTI 3D datasets demonstrate that our method outperforms the state-of-the-art competing approaches by wide margins. We also note that for \textit{hard} samples with heavier occlusions, our method demonstrates massive improvement.
In summary, the key contributions of this paper are as follows:
\begin{itemize}
	\item We design a novel 3D object detector using a monocular camera by capturing spatial relationships between paired objects, allowing largely improved accuracy on occluded objects.
	\item We propose an uncertainty-aware prediction module in 3D object detection, which is jointly optimized together with object-to-object distances.
  \item
	Experiments demonstrate that our method yields the best performance on KITTI 3D detection benchmark, by outperforming state-of-the-art competitors by wide margins.
\end{itemize}

\section{Related Work}
\label{sec:related}
In this section, we first review methods on monocular 3D object detection for autonomous driving. Related algorithms on object relationship and uncertainty estimation are also briefly discussed.

\vspace{2pt}\noindent\textbf{Monocular 3D Object Detection.}\hspace{5pt}
Monocular image is naturally of limited 3D information compared with multi-beam LiDAR or stereo vision.
Prior knowledge or auxiliary information are widely used for 3D object detection.
Mono3D \cite{chen_monocular_2016} focuses on the fact that 3D objects are on the ground plane.
Prior 3D shapes of vehicles are also leveraged to reconstruct the bounding box for autonomous driving \cite{murthy2017reconstructing}.
Deep MANTA \cite{chabot2017deep} predicts 3D object information utilizing key points and 3D CAD models.
SubCNN \cite{xiang_subcategory-aware_2017} learns viewpoint-dependent subcategories from 3D CAD models to capture both shape, viewpoint and occlusion patterns.
In \cite{barabanau_monocular_2019},  the network learns to estimate correspondences between detected 2D keypoints and 3D counterparts.
3D-RCNN \cite{kundu20183d} introduces an inverse-graphics framework for all object instances from an image. A differentiable Render-and-Compare loss allows 3D results to be learned through 2D information.
In \cite{ku_monocular_2019}, a sparse LiDAR scan is used in the training stage to generate training data, which removes the necessity of using inconvenient CAD dataset.
An alternative family of methods is to predict a stand-alone depth or disparity information of the monocular image at the first stage \cite{ma_accurate_2019, manhardt_roi_10d_2018, wang_pseudo-lidar_2018, xu_multilevel_2018}. Although they only require the monocular image at testing time, ground-truth depth information is still necessary for the model training.

Compared with the aforementioned works
in monocular 3D detection,  some algorithms consist of only the RGB image as input rather than relying on external data, network structures or  pre-trained models.
Deep3DBox \cite{mousavian_3d_2017} infers 3D information from a 2D bounding box considering the geometrical constraints of projection.
OFTNet \cite{roddick2018orthographic} presents a orthographic feature transform to map image-based features into an orthographic 3D space.
ROI-10D \cite{manhardt_roi_10d_2018} proposes a novel loss to properly measure the metric misalignment of boxes.
MonoGRNet \cite{qin_monogrnet_2018} predicts 3D object localizations from a monocular RGB image considering geometric reasoning in 2D projection and the unobserved depth dimension.
Current state-of-the-art results for monocular 3D object detection are from MonoDIS \cite{simonelli_disentangling_2019} and M3D-RPN \cite{brazil_m3d_rpn_2019}. Among them, MonoDIS \cite{simonelli_disentangling_2019} leverages a novel disentangling transformation for 2D and 3D detection losses, which simplifies the training dynamics.
M3D-RPN \cite{brazil_m3d_rpn_2019} reformulates the monocular 3D detection problem as a standalone 3D region proposal network.
However, all the object detectors mentioned above focus on predicting each individual object from the image. The spatial relationship among objects is not considered.
Our work is originally inspired by CenterNet \cite{zhou_objects_2019}, in which each object is identified by points.
Specifically, we model the geometric relationship between objects by using a single point similar to CenterNet, which is effectively the geometric center between them.
%

\begin{figure*}[!ht]
	\begin{center}
		\includegraphics[width=1.0\linewidth]{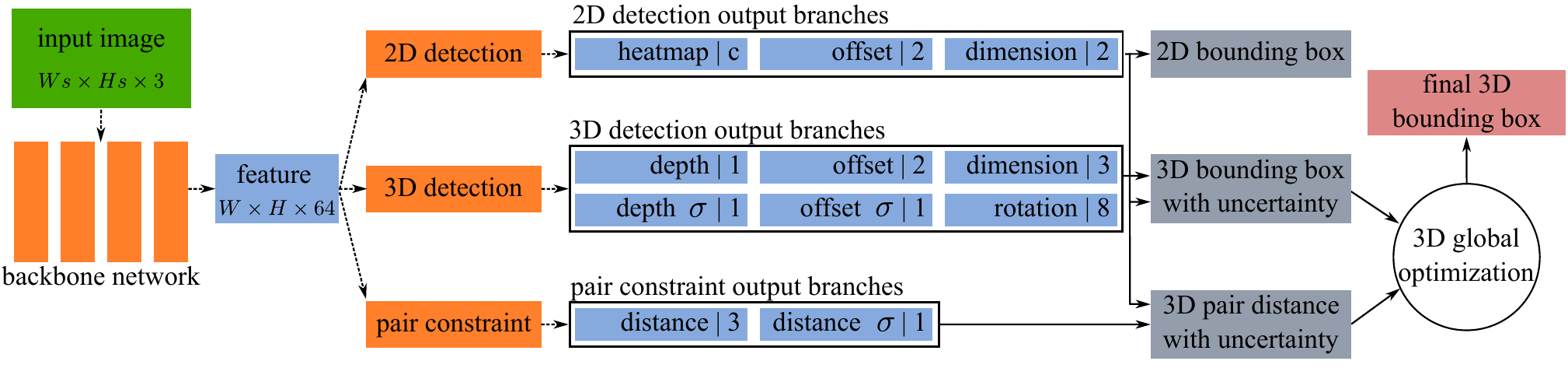}
	\end{center}
	\caption{Overview of our architecture. A monocular RGB image is taken as the input to the backbone network and trained with supervision. Eleven different prediction branches, with feature map as $W\times H \times m$, are divided into three parts: 2D detection, 3D detection and pair constraint prediction.
	The width and height of the output feature $(W, H)$ are as the same as the backbone output.
	Dash lines represent forward flows of the neural network. The heatmap and offset of 2D detection are also utilized to locate the 3D object center and the pairwise constraint keypoint.
	}
	\label{fig:overview}
\end{figure*}

\vspace{2pt}\noindent\textbf{Visual Relationship Detection.}\hspace{5pt}
Relationship plays an essential role for image understanding. To date, it is widely applied in image captioning.
Dai \textit{et al.} \cite{dai2017detecting} proposes a relational network to exploit the statistical dependencies between objects and their relationships.
MSDB \cite{Li_2017_ICCV} presents a multi-level scene description network to learn features of different semantic levels.
Yao \textit{et al.} \cite{Yao_2018_ECCV} proposes an attention-based encoder-decoder framework. through graph convolutional networks and long short-term memory (LSTM) for scene generation.
However, these methods are mainly for tackling the effects of visual relationships in representing and describing an image. They usually extract object proposals directly or show full trust for the predicted bounding boxes. By contrast, our method focuses 3D object detection, which is to refine the detection results based on spatial relationships. This is un-explored in existing work.

\vspace{2pt}\noindent\textbf{Uncertainty Estimation in object detection.}\hspace{5pt}
The computed object locations and pairwise 3D distances of our method are all predicted with uncertainties. This is inspired by the aleatoric uncertainty of deep neural networks \cite{gal2016uncertainty, kendall2017uncertainties}. Instead of fully trusting the results of deep neural networks, we can extract how uncertain the predictions. This is crucial for various perception and decision making tasks, especially for autonomous driving, where human lives may be endangered due to inappropriate choices. This concept has been applied in 3D Lidar object detection \cite{feng2018towards} and pedestrian localization \cite{bertoni2019monoloco}, where they mainly consider uncertainties as additional information for reference.
In \cite{wirges_capturing_2019}, uncertainty is used to approximate object hulls with bounded collision probability for subsequent trajectory planning tasks.
Gaussian-YOLO \cite{choi_gaussian_2019} significantly improves the detection results by predicting the localization uncertainty.
These approaches only use uncertainty to improve the training quality or to provide an additional reference. By contrast, we use uncertainty to weight the cost function for post-optimization, integrating the detection estimates and predicted uncertainties in global context optimization.

\section{Approach}
\label{sec:approach}

\subsection{Overview}

We adopt a one-stage architecture, which shares a similar structure with state-of-the-art anchor-free 2D object detectors \cite{tian_fcos_2019,zhou_objects_2019}.
As shown in Figure \ref{fig:overview}, it is composed of a backbone network and several task-specific dense prediction branches.
The backbone takes a monocular image $I$ with a size of $(Ws \times Hs)$ as input, and outputs the feature map with a size of $(W \times H \times 64)$, where $s$ is our backbone's down-sampling factor.
There are eleven output branches with a size of $W\times H \times m$, where $m$ means the channel of each output branch, as shown in Figure \ref{fig:overview}.
Eleven output branches are divided into three parts: three for 2D object detection, six for 3D object detection, and two for pairwise constraint prediction. We introduce each module in details as follows.



\begin{figure}[!t]
	\includegraphics[width=1.0\linewidth]{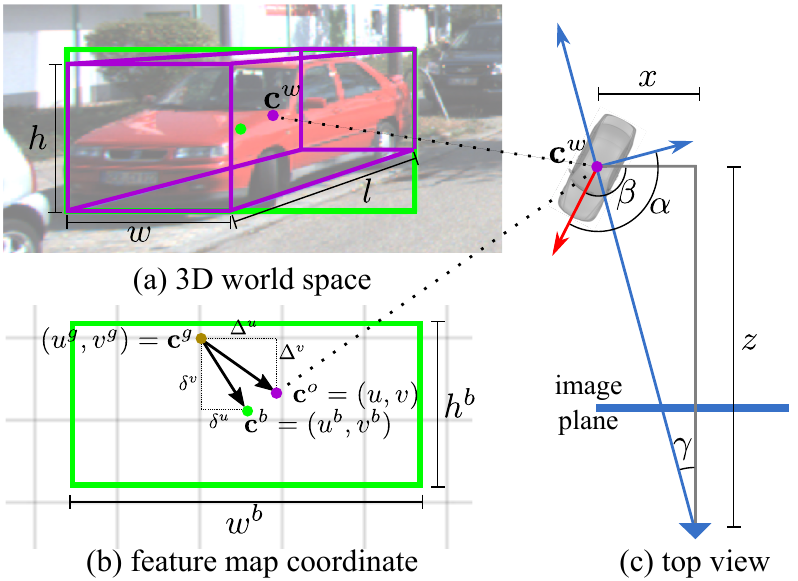}
	\caption{
	Visualization of notations for (a) 3D bounding box in world space, (b) locations of an object in the output feature map, and (c) orientation of the object from the top view. 3D dimensions are in meters, and all values in (b) are in the feature coordinate. The vertical distance $y$ is invisible and skipped in (c).
	}
	\label{fig:2d_3d_det}
\end{figure}

\subsection{2D Detection}

Our 2D detection module is derived from the CenterNet \cite{zhou_objects_2019} with three output branches. The heatmap with a size of ($W \times H \times c$) is used for keypoint localization and classification. Keypoint types include $c=3$ in KITTI3D object detection. Details about extracting the object location $\bc^g=(u^g, v^g)$ from the output heatmap can be referred in \cite{zhou_objects_2019}.
The other two branches, with two channels for each, output the size of the bounding box $(w^b, h^b)$ and the offset vector $(\delta^u, \delta^v)$ from the located keypoint $\bc^g$ to the bounding box center $\bc^b=(u^b, v^b)$ respectively. As shown in Figure \ref{fig:2d_3d_det}, those values are in units of the feature map coordinate.

\subsection{3D Detection}
The object center in world space is represented as $\bc^w=(x, y ,z)$. Its projection in the feature map is $\bc^o=(u, v)$ as shown in Figure \ref{fig:2d_3d_det}.
Similar to \cite{manhardt_roi_10d_2018, simonelli_disentangling_2019},
we predict its offset $(\Delta^u, \Delta^v)$ to the keypoint location $\bc^g$ and the depth $z$ in two separate branches. With the camera intrinsic matrix $\bK$, the derivation from predictions to the 3D center $\bc^w$ is as follows:
\begin{align}
    \bK &=
        \begin{bmatrix}
             f_x & 0 & a_x \\
            0 & f_y & a_y \\
            0 & 0 & 1
        \end{bmatrix}.\\
    \bc^w &= ( \frac{u^g+\Delta^u- a_x}{f_x}z, \frac{v^g+\Delta^v- a_y}{f_y}z, z) \label{equ:backpro}
\end{align}
Given the difficulty to regress depth directly, depth prediction branch outputs inverse depth $\hat{z}$ similar to \cite {eigen2014depth}, transforming the absolute depth by inverse sigmoid transformation $z=1/\sigma(\hat{z})-1$.
The dimension branch regresses the size $(w, h, l)$ of the object in meters directly.
The branches for depth, offset and dimensions in both 2D and 3D detection are trained with the L1 loss following \cite{zhou_objects_2019}.

As presented in Figure \ref{fig:2d_3d_det}, we estimate the object's local orientation $\alpha$ following \cite{mousavian_3d_2017} and \cite{zhou_objects_2019}. Compared to global orientation $\beta$ in the camera coordinate system, the local orientation accounts for the relative rotation of the object to the camera viewing angle $\gamma=\arctan(x/z)$. Therefore, using the local orientation is more meaningful when dealing with image features.
Similar to \cite{mousavian_3d_2017, zhou_objects_2019}, we represent the orientation using eight scalars, where the orientation branch is trained by $MultiBin$ loss.

\subsection{Pairwise Spatial Constraint}

\begin{figure}[!t]
  \centering
  \begin{subfigure}{0.4\columnwidth}
		\includegraphics[width=\linewidth]{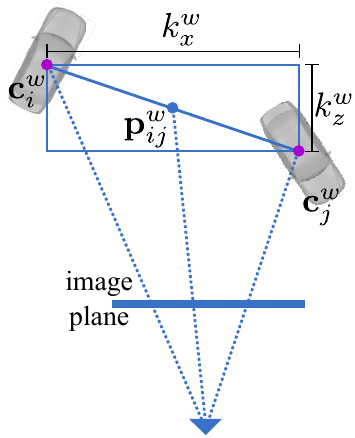}
		\caption{camera coordinate}
		\label{fig:cons_key_cam}
	\end{subfigure}
  \begin{subfigure}{0.4\columnwidth}
		\includegraphics[width=\linewidth]{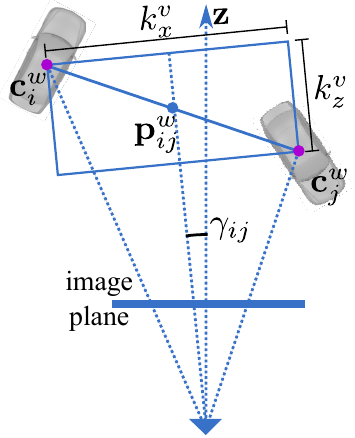}
		\caption{local coordinate}
		\label{fig:cons_key_view}
	\end{subfigure}
	\caption{
  Pairwise spatial constraint definition. $\bc^w_i$ and $\bc^w_j$ are centers of two 3D bounding boxes where $\bp^w_{ij}$ is their middle point. 3D distance in camera coordinate $\bk^w_{ij}$ and local coordinate $\bk^v_{ij}$ are shown in (a) and (b) respectively. The distance along $\mathbf{y}$ axis is skipped.
  }
	\label{fig:constraint_def}
\end{figure}

\label{sec:pair_select}
\begin{figure}
	\centering
	\begin{subfigure}{1.0\linewidth}
		\includegraphics[width=\linewidth]{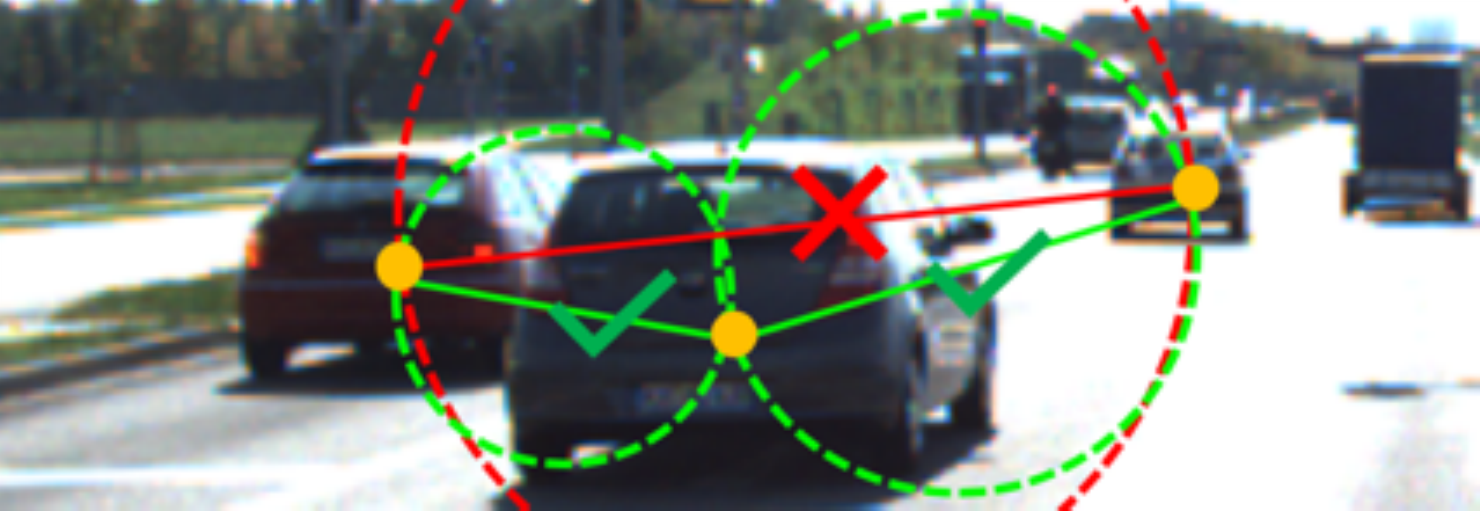}
		\caption{}
		\label{fig:pair_match}
	\end{subfigure}
	\begin{subfigure}{1.0\linewidth}
		\includegraphics[width=\linewidth]{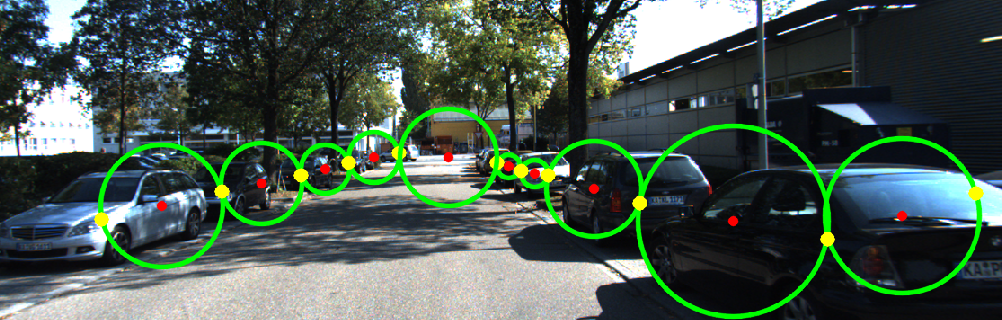}
		\caption{}
		\label{fig:pair_visual}
	\end{subfigure}
	\caption{
	\protect{Pair matching strategy for training and inference.}
	}
	\label{fig:pair_demo}
\end{figure}

In addition to the regular 2D and 3D detection pipelines, we propose a novel regression target, which is to estimate the pairwise geometric constraint among adjacent objects via a keypoint on the feature map.
Pair matching strategy for training and inference is shown in Figure \ref{fig:pair_match}.
For arbitrary sample pair, we define a range circle by setting the distance of their 2D bounding box centers as the diameter. This pair is neglected if it contains other object centers. Figure \ref{fig:pair_visual} shows an example image with all effective sample pairs.

\begin{figure}[!t]
  \centering
  \begin{subfigure}{0.42\columnwidth}
		\includegraphics[width=\linewidth]{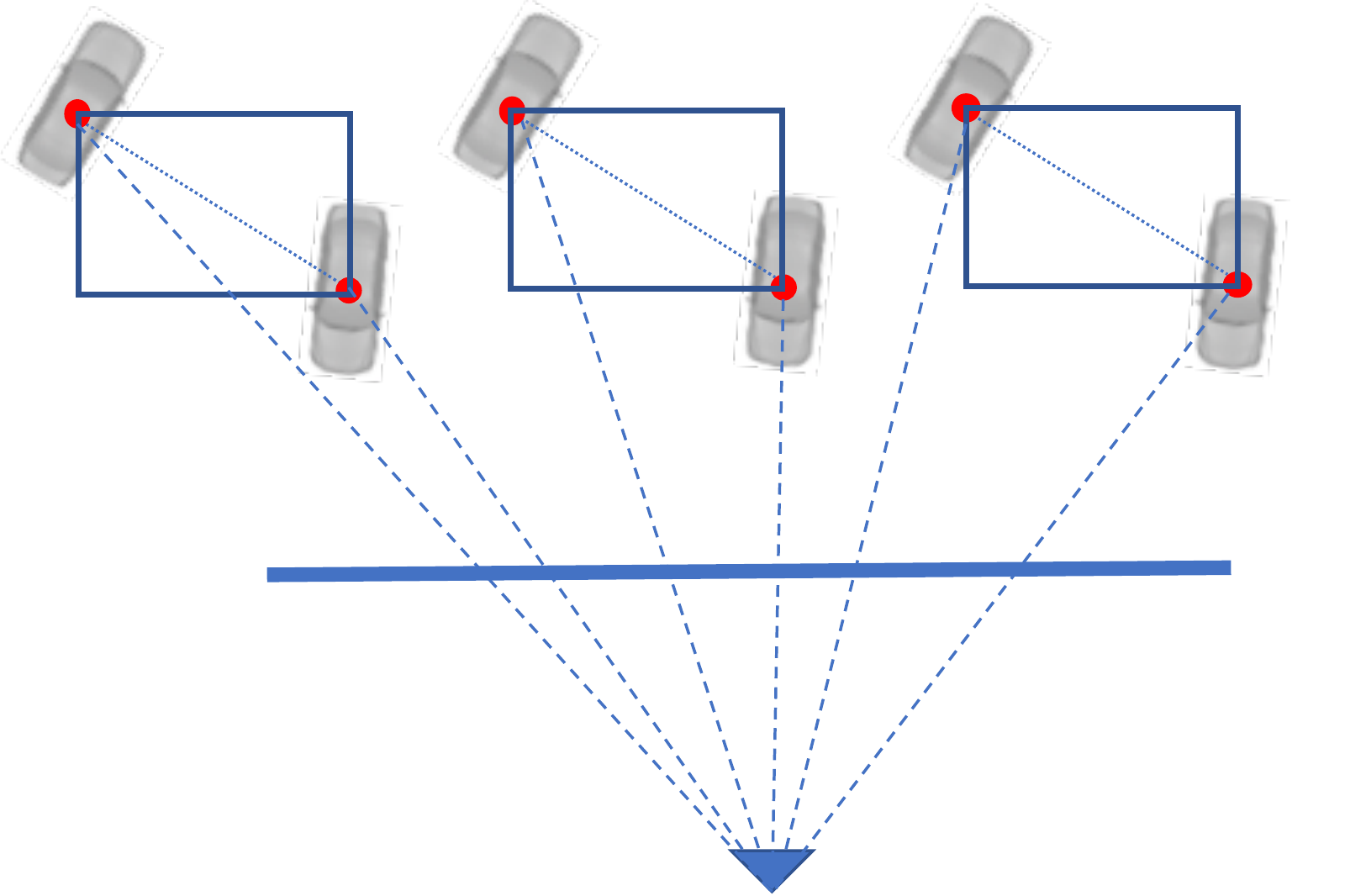}
		\caption{camera coordinate}
		\label{fig:cons_v_cam}
	\end{subfigure}
  \begin{subfigure}{0.435\columnwidth}
		\includegraphics[width=\linewidth]{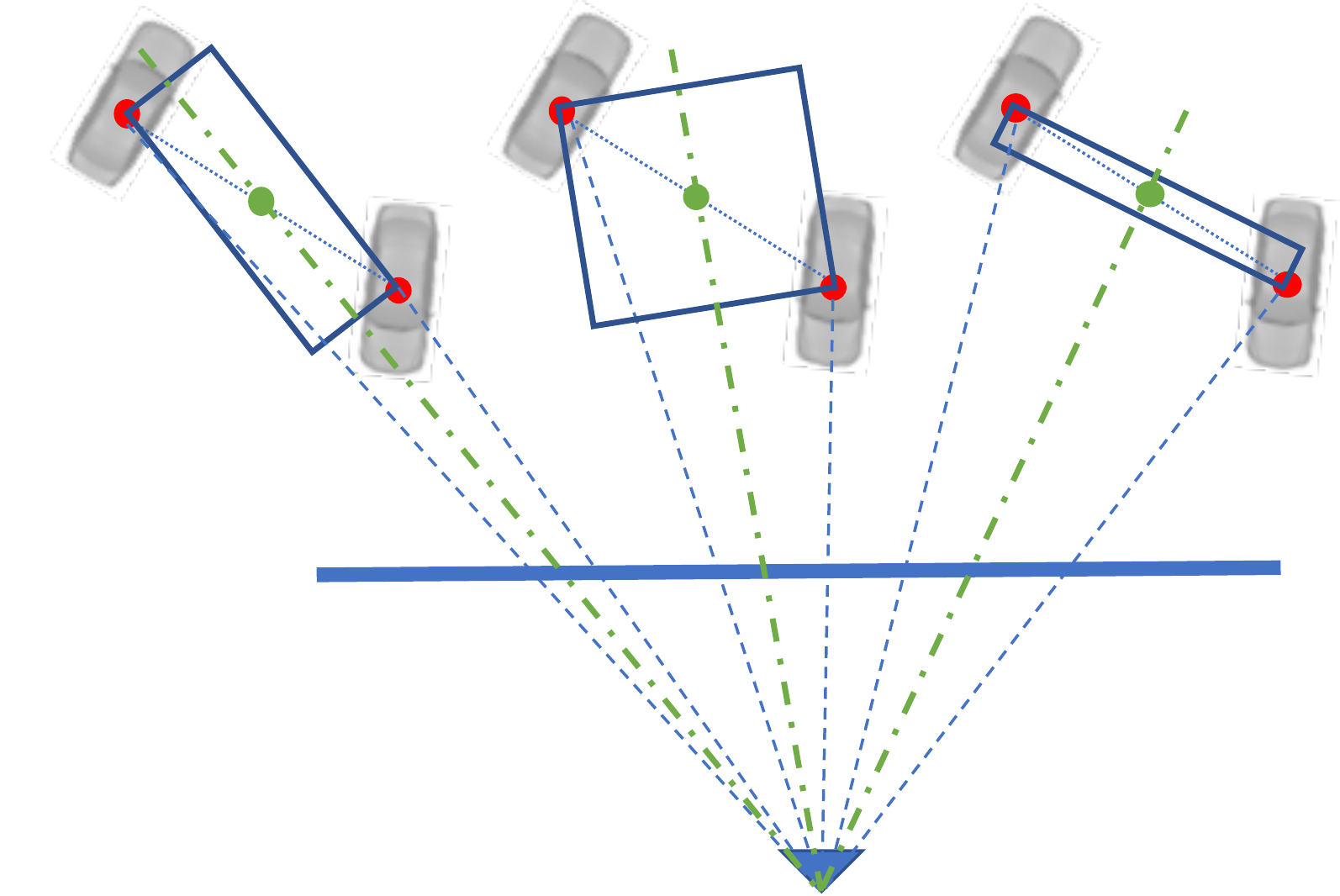}
		\caption{local coordinate}
		\label{fig:cons_v_loc}
	\end{subfigure}
	\caption{
  The same pairwise spatial constraint in camera and local coordinates from various viewing angles. The spatial constraint in camera coordinate is invariant among different view angles. Considering the different projected form of the car, we use the 3D absolute distance in local coordinate as the regression target of spatial constraint.
  }
	\label{fig:constraint_global_vs_local}
\end{figure}

\begin{figure*}[!ht]
  \centering
  \begin{subfigure}{0.23\linewidth}
		\includegraphics[width=\linewidth]{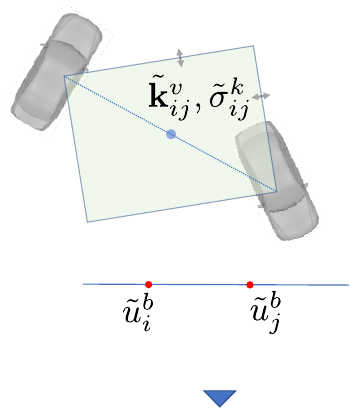}
		\caption{pair constraint prediction}
		\label{fig:optim_pair}
	\end{subfigure}
  \begin{subfigure}{0.27\linewidth}
		\includegraphics[width=\linewidth]{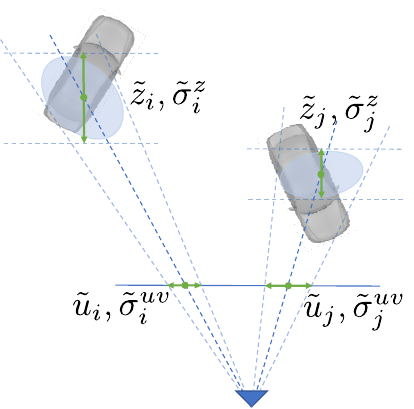}
		\caption{object location prediction}
		\label{fig:optim_single}
	\end{subfigure}
  \begin{subfigure}{0.23\linewidth}
		\includegraphics[width=\linewidth]{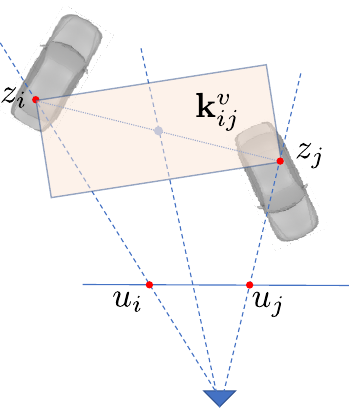}
		\caption{variables of optimization}
		\label{fig:optim_target}
	\end{subfigure}
  \begin{subfigure}{0.23\linewidth}
		\includegraphics[width=\linewidth]{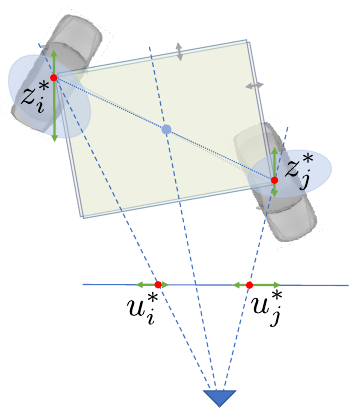}
		\caption{optimized results}
	\end{subfigure}
	\caption{
  Visualization of optimization for an example pair including. In (a), The predicted pairwise constraint $\tilde{\bk}^v_{ij}$ and its uncertainty $\tilde{\sigma}^k_{ij}$ is located by predicted 2D bounding box centers $(\tilde{u}^b_i, \tilde{v}^b_i )$ and $(\tilde{u}^b_j, \tilde{v}^b_j)$ on the feature map.
  The 3D prediction results (green points) are shown in (b). All uncertainties are represented as arrows to show a confidence range.
  We show variables in (c) for this optimization function as red points. The final optimized results are presented in (d). Our method is mainly supposed to work for occluded samples. The relatively long distance among the paired cars is for simplicity in visualization. Properties along $v$ direction is skipped.
  }
     \label{fig:optimization}
 \end{figure*}

Given a selected pair of objects, their 3D centers in world space are
$\mathbf{c}^w_i=(x_i,y_i,z_i)$ and $\mathbf{c}^w_j=(x_j,y_j,z_j)$
and their 2D bounding box centers on the feature map are
$\bc^b_i = (u^b_i, v^b_i)$ and $\bc^b_j = (u^b_j, v^b_j)$ .
The pairwise constraint keypoint locates on the feature map as $\bp^b_{ij}=(\bc^b_i+\bc^b_j)/2$.
The regression target for the related keypoint is the 3D distance of these two objects.
We first locate the middle point $\bp^w_{ij}=(\bc^w_i + \bc^w_j)/2 = (p^w_x, p^w_y, p^w_z)_{ij}$ in 3D space. Then, the 3D absolute distance $\bk^v_{ij} = (k^v_x, k^v_y, k^v_z)_{ij}$ along the view point direction, as shown in Figure \ref{fig:cons_key_view}, are taken as the regression target which is the distance branch of the pair constraint output in Figure \ref{fig:overview}.
Notice that $\bp^b$ is not the projected point of $\bp^w$ on the feature map, like $\bc^w$ and $\bc^b$ in Figure \ref{fig:2d_3d_det}.

For training, $\bk^v_{ij}$ can be easily collected through the groundtruth 3D object centers from the training data as:
\begin{equation}
	\bk^v_{ij}=\overrightarrow{\left| \bR(\gamma_{ij})\bk^w_{ij} \right|},
\end{equation}
where $\overrightarrow{ \left|~\cdot~\right| }$ means extract absolute value of each entry in the vector. $\bk^w_{ij} = \bc^w_i - \bc^w_j$ is the 3D distance in camera coordinate, $\gamma_{ij}=\arctan(p^w_x/p^w_z)$ is the view direction of their middle point $\bp^w_{ij}$, and $\bR(\gamma_{ij})$ is its rotation matrix along the $\bY$ axis as
\begin{equation}
	\bR(\gamma_{ij})=
		\begin{bmatrix}
			\cos(\gamma_{ij}) & 0 & -\sin(\gamma_{ij}) \\
			0 & 1 & 0 \\
			\sin(\gamma_{ij}) & 0 & \cos(\gamma_{ij})
		\end{bmatrix}.
\end{equation}

The 3D distance $\bk^w$ in camera coordinate is not considered because it is invariant from different view angles, as shown in Figure \ref{fig:cons_v_cam}. As in estimation of the orientation $\gamma$, 3D absolute distance $\bk^v$ in the local coordinate of $\bp^w$ is more meaningful considering the appearance change through viewing angles.

In inference, we first estimate objects' 2D locations and extract pairwise constraint keypoint located in the middle of predicted 2D bounding box centers. The predicted $\tilde{\bk}^v$ is extracted in the dense feature map of the distance branch based on the keypoint location. We do not consider offsets for this constraint keypoint both in training and reference, and round the middle point $\bp^b_{ij}$ of paired objects' 2D centers to the nearest grid point on the feature map directly.

\subsection{Uncertainty}

Following the heteroscedastic aleatoric uncertainty setup in \cite{kendall2017uncertainties, kendall2019geometry}, we represent a regression task with L1 loss as
\begin{align}
[\tilde{\by}, \tilde{\sigma}] &= f^{\theta}(\bx), \\
L(\theta) &= \frac{ \sqrt{2}}{\tilde\sigma} \| \by - \tilde\by \| + \log \tilde\sigma.
\end{align}
Here, $\bx$ is the input data, $\by$ and $\tilde\by$ are the groundtruth regression target and the predicted result. $\tilde\sigma$ is another output of the model and can represent the observation noise of the data $\bx$. $\theta$ is the weight of the regression model.


As mentioned in \cite{kendall2017uncertainties}, aleatoric uncertainty $\tilde\sigma(\bx)$
makes the loss more robust to noisy input in a regression task.
In this paper, we add three uncertainty branches as shown as $\sigma$ blocks in Figure \ref{fig:overview} for the depth prediction $\sigma^z$,  3D center offset $\sigma^{uv}$ and pairwise distance $\sigma^k$ respectively. They are mainly used to weight the error terms as presented in Section \ref{sec:g_optim}.

\subsection{Spatial Constraint Optimization}
\label{sec:g_optim}

 As the main contribution of this paper, we propose a post-optimization process from a graph perspective. Suppose that in one image, the network outputs $N$ effective objects, and there are $M$ pair constraints among them based on the strategy in Section \ref{sec:pair_select}.
 Those paired objects are regarded as vertices ${\{\bxi_i\}^{N^{\mathcal{G}}}_{i=1}}$ with size of $N^{\mathcal{G}}$ and the $M$ paired constraints are regarded as edges of the graph. Each vertex may connect multiple neighbors. Predicted objects not connected by other vertices are not updated anymore in the post-optimization.
 The proposed spatial constraint optimization is formulated as a nonlinear least square problem as
 \begin{equation}
         \mathop{\arg\min}_{(u_i, v_i, z_i)^{N^{\mathcal{G}}}_{i=1}} \be^T\bW\be,
 \end{equation}
 where $\be$ is the error vector and $\bW$ is the weight matrix for different errors.
 $\bW$ is a diagonal matrix with dimension $3N^\mG+3M$. For each vertex $\bxi_i$, there are three variables $(u_i, v_i, z_i)$, which are the projected center $(u_i, v_i)$ of the 3D bounding box on the feature map and the depth $z_i$ as shown in Figure \ref{fig:2d_3d_det}.
 We introduce each minimization term in the following.

 \vspace{2pt}\noindent\textbf{Pairwise Constraint Error}\hspace{5pt}
 For each pairwise constraint connecting $\bxi_i$ and $\bxi_j$, there are three error terms $(\be^x_{ij}, \be^y_{ij}, \be^z_{ij})$ measuring the inconsistency between network estimated 3D distance $\tilde{\bk}^v_{ij}$ and the distance $\bk^v_{ij}$ obtained by 3D locations $\bc^w_i$ and $\bc^w_j$ of the two associated objects.
  $\bc^w_i$ and $\bc^w_j$ can be represented by variables $(u_i, v_i, z_i)$,
 $(u_j, v_j, z_j)$ and the known intrinsic matrix through Equation \ref{equ:backpro}.
 Thus, error terms $(\be^x_{ij}, \be^y_{ij}, \be^z_{ij})$ are the absolute difference between $\tilde{\bk}^v_{ij}$ and $\bk^v_{ij}$ along three axis as following.
 \begin{align}
     \bk^v_{ij} & =\overrightarrow{\left| \bR(\gamma_{ij}) (\bc^w_i - \bc^w_j)  \right|} \\
     (\be^x_{ij}, \be^y_{ij}, \be^z_{ij})^T& =\overrightarrow{\left| \tilde{\bk}^v_{ij} - \bk^v_{ij} \right|}
 \end{align}
 \vspace{2pt}\noindent\textbf{Object Location Error}
 For each vertex $\bxi_i$, there are three error terms $(\be^u_i, \be^v_i, \be^z_i)$ to regularize the optimization variables with the predicted values from the network.
 We use this term to constraint the deviation between network estimated object location and the optimized location as follows.
 \begin{align}
     \be^u_i &=\left| \tilde{u}^g_i + \tilde{\Delta^u_i} - u_i \right| \\
     \be^v_i &=\left| \tilde{v}^g_i + \tilde{\Delta^v_i} - v_i \right| \\
     \be^z_i &=\left| \tilde{z}_i - z_i \right|
 \end{align}

 \vspace{2pt}\noindent\textbf{Weight Matrix}\hspace{5pt}
 The weight matrix $\bW$ is constructed by the uncertainty output $\tilde{\sigma}$ of the network. The weight of the error is higher when the uncertainty is lower, which means we have more confidence in the predicted output.
 Thus, we use $1/\tilde{\sigma}$ as the element of $\bW$.
 For pairwise inconsistency, the weights for the three error terms $(\be^x_{ij}, \be^y_{ij}, \be^z_{ij})$  are the same as the predicted $1/\tilde{\sigma}_{ij}$ as shown in Figure \ref{fig:optim_pair}.
 For object location error, the weight is $1/\tilde{\sigma}^z_{i}$ for depth error $\be^z_i$ and $1/\tilde{\sigma}^{uv}_{i}$ for both $\be^u_i$ and $\be^v_i$ as shown in Figure \ref{fig:optim_single}.
 We visualize an example pair for the spatial constraint optimization in Figure \ref{fig:optimization}. Uncertainties give us confidence ranges to tune variables so that both the pairwise constraint error and
 the object location error can be jointly minimized.
 We use g2o \cite{kummerle2011g} to conduct this graph optimization structure during implementation.

\section{Implementation}
\label{sec:implementation}

We conduct experiments on the challenge KITTI 3D object detection dataset \cite{Geiger2012CVPR}.
It is split to 3712 training samples and 3769 validation samples as \cite{chen_3d_2015}.
Samples are labeled from \textit{Easy}, \textit{Moderate}, to \textit{Hard} according to its condition of truncation, occlusions and bounding box height. Table \ref{tab:pair_count} shows counts of groundtruth pairwise constraints through the proposed pair matching strategy from all the training samples.

\begin{table}[!ht]
		\centering
		\renewcommand\arraystretch{1.0}
		\renewcommand{\tabcolsep}{3pt}
		\begin{tabular}{c|c|c|c}
			\hline
			Count&  object &  pair & paired object \\
			\hline
			Car & 14357 & 11110 &13620 \\
			Pedestrian & 2207 & 1187 &1614 \\
			Cyclist & 734 & 219 &371 \\
			\hline
		\end{tabular}
	\caption{Count of objects, pairs and paired objects of each category in the KITTI training set.}
	\label{tab:pair_count}
\end{table}


\subsection{Training}
We adopt the modified DLA-34 \cite{yu2018deep} as our backbone.
The resolution of the input image is set to $380 \times 1280$.
The feature map of the backbone output is with a size of $96 \times 320 \times 64$.
Each of the eleven output branches connects the backbone feature with two additional convolution layers with sizes of $3 \times 3 \times 256$ and $1 \times 1  \times m$, where $m$ is the feature channel of the related output branch.
Convolution layers connecting output branches maintain the same feature width and height. Thus, the feature size of each output branch is $96 \times 320 \times m$.

We train the whole network in an end-to-end manner for 70 epochs with a batchsize of 32 on four GPUs simultaneously.
The initial learning rate is 1.25e-4, dropped by multiplying 0.1 both at 45 and 60 epochs.
It is trained with Adam optimizer with weight decay as 1e-5. We conduct different data augmentation strategies during training, as random cropping and scaling for 2D detection, and random horizontal flipping for both 3D detection and pairwise constraints prediction.

\begin{table*}[!ht]
		\centering
		\renewcommand\arraystretch{1.0}
		\renewcommand{\tabcolsep}{3pt}
		\begin{tabular}{l|ccc|ccc|ccc|ccc|c }
			\hline
			\multirow{2}{*}{Methods}& \multicolumn{3}{c|}{$AP_{bv}$ IoU$\geq$0.5} & \multicolumn{3}{c|}{$AP_{3D}$ IoU$\geq$0.5} & \multicolumn{3}{c|}{$AP_{bv}$ IoU$\geq$0.7} & \multicolumn{3}{c|}{$AP_{3D}$ IoU$\geq$0.7} & RT \\
			\cline{2-13}
			& E & M & H  & E & M & H  & E & M & H  & E & M & H & (ms)\\
			\hline
			CenterNet\cite{zhou_objects_2019}*	&34.36	&27.91	&24.65	&20.00	&17.50	&15.57	&3.46	&3.31	&3.21	&0.60	&0.66	&0.77	& \textbf{45} \\
			MonoDIS\cite{simonelli_disentangling_2019} & - & - & - & - & - & - & 18.45 & 12.58 & 10.66 & 11.06 & 7.60 & 6.37 & -\\
			MonoGRNet\cite{qin_monogrnet_2018}* & 52.13 & 35.99 & 28.72 & 47.59 & 32.28 & 25.50 & 19.72 & 12.81 & 10.15 & 11.90 & 7.56 & 5.76 & 60\\ 
			M3D-RPN\cite{brazil_m3d_rpn_2019}*  & 53.35 & 39.60 & 31.76 & 48.53 & 35.94 & 28.59 & 20.85 & 15.62 & 11.88 & 14.53 & 11.07 & 8.65 & 161\\ 
			\hline
			Baseline & 53.06 & 38.51 & 32.56 & 47.63 & 33.19 & 28.68 & 19.83 & 12.84 & 10.42 & 13.06 &  7.81 &  6.49 & 47 \\
			$+\sigma^{z}+\sigma^{uv}$ & 59.22 & 46.90 & 41.38 & 53.44 & 41.46 & 36.28 & 21.71 & 17.39 & 15.10 & 14.75 & 11.42 &  9.76& 50 \\
			MonoPair & \textbf{61.06} & \textbf{47.63} & \textbf{41.92} & \textbf{55.38} & \textbf{42.39} & \textbf{37.99} & \textbf{24.12} & \textbf{18.17} & \textbf{15.76} & \textbf{16.28} & \textbf{12.30} & \textbf{10.42} & 57  \\
			\hline
		\end{tabular}
	\caption{$AP_{40}$ scores on KITTI3D validation set for car. * indicates that the value is extracted by ourselves from the public pretrained model or results provided by related paper author. E, M and H represent \textit{Easy}, \textit{Moderate} and \textit{Hard} samples.}
	\label{tab:ap40}
\end{table*}

\begin{table*}[!ht]
		\centering
		\renewcommand\arraystretch{1.0}
		\renewcommand{\tabcolsep}{3pt}
		\begin{tabular}{c|ccc|ccc|ccc|ccc}
			\hline
			\multirow{2}{*}{Methods}&  \multicolumn{3}{c|}{$AP_{2D}$} &  \multicolumn{3}{c|}{$AOS$} &  \multicolumn{3}{c|}{$AP_{bv}$} & \multicolumn{3}{c}{$AP_{3D}$} \\
			\cline{2-13}
			& E & M & H  & E & M & H  & E & M & H  & E & M & H  \\
			\hline
			MonoGRNet\cite{qin_monogrnet_2018} & 88.65 & 77.94 & 63.31 & - & - & - & 18.19 & 11.17 & 8.73 & 9.61 & 5.74 & 4.25 \\
			MonoDIS\cite{simonelli_disentangling_2019} &94.61 &89.15 &78.37 &- &- &- &17.23 &13.19 &11.12 &10.37 &7.94 &6.40  \\
			M3D-RPN\cite{brazil_m3d_rpn_2019} &89.04 &85.08 &69.26 &88.38 &82.81 &67.08 & $\textbf{21.02}$ &13.67 &10.23 & $\textbf{14.76}$ &9.71 &7.42  \\
			\hline
			MonoPair & \textbf{96.61} & \textbf{93.55} & \textbf{83.55} & \textbf{91.65} & \textbf{86.11} & \textbf{76.45} & 19.28 & \textbf{14.83} & \textbf{12.89} & 13.04 & \textbf{9.99} & \textbf{8.65} \\
			\hline
		\end{tabular}
	\caption{$AP_{40}$ scores on  KITTI3D test set for car referred from the KITTI benchmark website.}
	\label{tab:ap40_test_full}
\end{table*}

\subsection{Evaluation}

Following \cite{simonelli_disentangling_2019}, we use 40-point interpolated average precision metric $AP_{40}$ that averaging precision results on 40 recall positions instead of 0. The previous metric $AP_{11}$ of KITTI3D average precision on 11 recall positions, which may trigger bias to some extent.
The precision is evaluated at both the bird-eye view 2D box $AP_{bv}$ and the 3D bounding box $AP_{3D}$ in world space.
We report average precision with intersection over union (IoU) using both 0.5 and 0.7 as thresholds.

For the evaluation and ablation study, we show experimental results from three different setups. \textbf{Baseline} is derived from CenterNet \cite{zhou_objects_2019} with an additional output branch to represent the offset of the 3D projected center to the located keypoint.
\textbf{$+\sigma^{z}+\sigma^{uv}$} adds two uncertainty prediction branches on \textbf{Baseline} which consists of all the three 2D detection branches and six 3D detection branches as shown in Figure \ref{fig:overview}. \textbf{MonoPair} is the final proposed method integrating the eleven prediction branches and the pairwise spatial constraint optimization.



\begin{table}[!t]
		\centering
		\renewcommand\arraystretch{1.0}
		\renewcommand{\tabcolsep}{3pt}
		\begin{tabular}{c|c|ccc|ccc}
			\hline
			\multirow{2}{*}{Cat}& \multirow{2}{*}{Method}&    \multicolumn{3}{c|}{$AP_{bv}$} & \multicolumn{3}{c}{$AP_{3D}$} \\
			\cline{3-8}
			& & E & M & H  & E & M & H  \\
			\hline
			\multirow{2}{*}{Ped}&M3D-RPN\cite{brazil_m3d_rpn_2019} &5.65 &4.05 &3.29 &4.92 &3.48 &2.94  \\
			\cline{2-8}

			& MonoPair & 10.99 & 7.04  & 6.29 & 10.02 & 6.68  & 5.53 \\

		\hline
		\multirow{2}{*}{Cyc}&M3D-RPN\cite{brazil_m3d_rpn_2019} &1.25 &0.81 &0.78 &0.94 &0.65 &0.47   \\
		\cline{2-8}

		& MonoPair & 4.76 & 2.87  & 2.42 & 3.79 & 2.12  & 1.83 \\
		\hline
	\end{tabular}
	\caption{$AP_{40}$ scores on pedestrian and cyclist samples from the KITTI3D test set at 0.7 IoU threshold. It can be referred from the KITTI benchmark website.}
	\label{tab:ap40_test_ped_cyc}
\end{table}

\section{Experimental Results}
\subsection{Quantitative and Qualitative Results}

We first show the performance of our proposed MonoPair on KITTI3D validation set for car, compared with other state-of-the-art (SOTA) monocular 3D detectors including MonoDIS \cite{simonelli_disentangling_2019}, MonoGRNet \cite{qin_monogrnet_2018} and M3D-RPN \cite{brazil_m3d_rpn_2019} in Table \ref{tab:ap40}.
Since MonoGRNet and M3D-RPN have not published their results through $AP_{40}$, we evaluate the related values through their published detection results or models.

As shown in Table \ref{tab:ap40}, although our baseline is only comparable or a little worse than SOTA detector M3D-RPN, MonoPair outperforms all the other detectors mostly by a large margin, especially for \textit{hard} samples with augmentations from the uncertainty and the pairwise spatial constraint.
Table \ref{tab:ap40_test_full} shows results of our MonoPair on the KITTI3D test set for car. From the KITTI 3D object detection benchmark\footnote{http://www.cvlibs.net/datasets/kitti/eval\_object.php?obj\_benchmark=3d}, we achieve the highest score for \textit{Moderate} samples and rank at the first place among those 3D monocular object detectors without using additional information. $AP_{2D}$ and $AOS$ are metrics for 2D object detection and orientation estimations following the benchmark. Apart from the \textit{Easy} result of $AP_bv$ and $AP_{3D}$, our method outperforms M3D-RPN for a large margin, especially for \textit{Hard} samples. It proves the effects of the proposed pairwise constraint optimization targeting for highly occluded samples.

\begin{figure*}[!ht]
	\begin{center}
		\hspace{0cm}
		\includegraphics[width=0.9\linewidth]{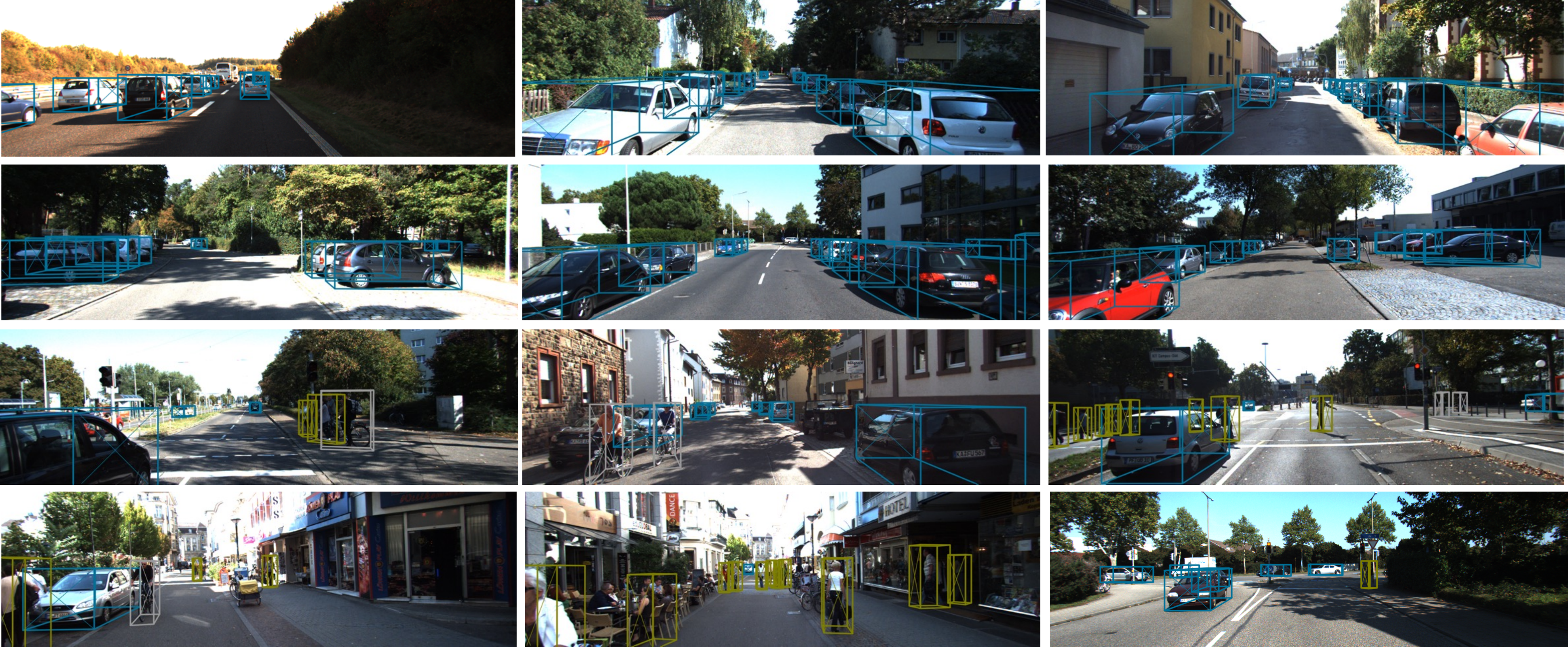}
	\end{center}
	\caption{Qualitative results in KITTI validation set. Cyan, yellow and grey mean predictions of car, pedestrian and cyclist.}
	\label{fig:qualitive}
\end{figure*}

We show the pedestrian and cyclist detection results on the KITTI test set in Table \ref{tab:ap40_test_ped_cyc}.
Because MonoDIS \cite{simonelli_disentangling_2019} and MonoGRNet \cite{qin_monogrnet_2018} do not report their performance on pedestrian and cyclist categories, we only compare our method with M3D-RPN \cite{brazil_m3d_rpn_2019}. It presents a significant improvement from our MonoPair. Even though the relatively few training samples of pedestrian and cyclist, the proposed pairwise spatial constraint goes much deeper by utilizing object relationships compared with target-independent detectors.

Besides, compared with those methods relying on time-consuming region proposal network \cite{brazil_m3d_rpn_2019, simonelli_disentangling_2019}, our one-stage anchor-free detector is more than two times faster on an Nvidia GTX 1080 Ti. It can perform inference in real-time as 57 ms per image, as shown in Table \ref{tab:ap40}.

%
%

\subsection{Ablation Study}
\begin{table}[!t]
		\centering
		\renewcommand\arraystretch{1.0}
		\renewcommand{\tabcolsep}{3pt}
		\begin{tabular}{c|cc|cc}
			\hline
			\multirow{2}{*}{Uncertainty}&  \multicolumn{2}{c|}{ IoU$\geq$0.5} & \multicolumn{2}{c}{ IoU$\geq$0.7} \\
			\cline{2-5}
			& $AP_{bv}$ & $AP_{3D}$ & $AP_{bv}$  & $AP_{3D}$  \\
			\hline
			Baseline & 38.51 & 33.19 & 12.84 & 7.81 \\
			$+\sigma^{uv}$ & 42.79 & 38.75 & 14.38 &  8.96 \\
			$+\sigma^z$ & 45.09 & 40.46 & 15.79 & 10.15 \\
			$+\sigma^z+\sigma^{uv}$ & 46.90 & 41.46 & 17.39 & 11.42 \\
			\hline
		\end{tabular}
	\caption{Ablation study for different uncertainty terms.}
	\label{tab:ap40_dv_cv}
\end{table}
%

\begin{table}[!t]
		\centering
		\renewcommand\arraystretch{1.0}
		\renewcommand{\tabcolsep}{3pt}
		\begin{tabular}{c|c|cc|cc}
			\hline
			\multirow{2}{*}{pairs}&\multirow{2}{*}{images}&  \multicolumn{2}{c|}{$AP_{bv}$} & \multicolumn{2}{c}{$AP_{3D}$} \\
			\cline{3-6}
			& & Uncert.  & MonoPair & Uncert.  & MonoPair  \\
			\hline
			0-1 & 1404 & 10.40 & 10.44 & 5.41 &6.02 \\
			2-4 & 1176 & 13.25 & 14.00 & 8.46 &8.97 \\
			5-8 & 887 & 20.45 & 22.32 & 14.63 &15.54 \\
			9- & 302 & 25.49 & 25.87 & 17.98 &1894 \\
			\hline
		\end{tabular}
	\caption{Ablation study for improvements among different pair counts through 0.7 IoU.}
	\label{tab:ap40_min_max_pair}
\end{table}

We conduct two ablation studies for different uncertain terms and the count of pairwise constraints both on KITTI3D validation set through ${AP_{40}}$. We only show results from \textit{Moderate} samples here.

For uncertainty study, except the \textbf{Baseline} and \textbf{$+\sigma^{z}+\sigma^{uv}$} setups mentioned above, we add \textbf{$\sigma^z$} and \textbf{$\sigma^{uv}$} methods by only predict the depth or projected offset uncertainty based on the \textbf{Baseline}.
From Table \ref{tab:ap40_dv_cv}, uncertainties prediction from both depth and offset show considerable development above the baseline, where the improvement from depth is larger. The results match the fact that depth prediction is a much more challenging task and it can benefit more from the uncertainty term.
It proves the necessity of imposing uncertainties for 3D object prediction, which is rarely considered by previous detectors.

In terms of the pairwise constraint, we divide the validation set to different parts based on the count of groundtruth pairwise constraints. The \textbf{Uncert.} in Table \ref{tab:ap40_min_max_pair} represents \textbf{$+\sigma^{z}+\sigma^{uv}$} for simplicity. By checking both the $AP_{bv}$ and ${AP_{3D}}$ in Table \ref{tab:ap40_min_max_pair}, the third group with 5 to 8 pairs shows higher average precision improvement. A possible explanation is that fewer pairs may not provide enough constraints, and more pairs may increase the complexity of the optimization.

Also, to prove the utilization of using uncertainties to weigh related errors, we tried various strategies for weight matrix designing, for example, giving more confidence for objects closed to the camera or setting the weight matrix as identity. However, none of those strategies showed improvements in the detection performance. On the other hand, the baseline is easily dropped to be worse because of coarse post-optimization. It shows that setting the weight matrix of the proposed spatial constraint optimization is nontrivial. And uncertainties, besides its original function to enhance network training, is naturally a meaningful choice for weights of different error terms.


%
%

\section{Conclusions}

We proposed a novel post-optimization method for 3D object detection with uncertainty-aware training from a monocular camera. By imposing aleatoric uncertainties into the network and considering spatial relationships for objects, our method has achieved the state-of-the-art performance on KITTI 3D object detection benchmark using a monocular camera without additional information.
By exploring the spatial constraints of object pairs, we observed the enormous potential of geometric relationships in object detection, which was rarely considered before. For future work, finding spatial relationships across object categories and innovating pair matching strategies would be exciting next steps.

%

\clearpage

{\small
\bibliographystyle{ieee_fullname}
\bibliography{3dod}
}

\end{document}